\documentclass[letterpaper]{article} 
\usepackage{aaai25}  
\usepackage{times}  
\usepackage{helvet}  
\usepackage{courier}  
\usepackage[hyphens]{url}  
\usepackage{graphicx} 
\usepackage{natbib}  
\usepackage{caption} 
\usepackage{algorithm}
\usepackage{algorithmic}
\usepackage{multirow}
\usepackage{amsmath}
\usepackage{makecell}
\usepackage{amssymb}
\usepackage{colortbl}
\usepackage{booktabs}
\usepackage{soul}
\usepackage{subcaption} 
\usepackage{pifont}
\usepackage{capt-of}
\usepackage{subfig}
\usepackage{listings}
\DeclareCaptionStyle{ruled}{labelfont=normalfont,labelsep=colon,strut=off}
\floatstyle{ruled}
\newfloat{listing}{tb}{lst}{}
\floatname{listing}{Listing}

\title{DynASyn: Multi-Subject Personalization Enabling Dynamic Action Synthesis}

\author {
    Yongjin Choi,
    Chanhun Park,
    Seung Jun Baek\thanks{Corresponding Author}
}
\affiliations {
    Department of Computer Science, Korea University, Seoul, Korea\\
    \{dydwls8445, cksgns2010, sjbaek\}@korea.ac.kr
}

\begin{document}

\maketitle


\begin{abstract}
Recent advances in text-to-image diffusion models spurred research on personalization, i.e., a customized image synthesis, of subjects within reference images. Although existing personalization methods are able to alter the subjects' positions or to personalize multiple subjects simultaneously, they often struggle to modify the behaviors of subjects or their dynamic interactions. The difficulty is attributable to overfitting to reference images, which worsens if only a single reference image is available. We propose DynASyn, an effective multi-subject personalization from a single reference image addressing these challenges. DynASyn preserves the subject identity in the personalization process by aligning concept-based priors with subject appearances and actions. This is achieved by regularizing the attention maps between the subject token and images through concept-based priors. 
In addition, we propose concept-based prompt-and-image augmentation for an enhanced trade-off between identity preservation and action diversity. We adopt an SDE-based editing guided by augmented prompts to generate diverse appearances and actions while maintaining identity consistency in the augmented images. 
Experiments show that DynASyn is capable of synthesizing highly realistic images of subjects with novel contexts and dynamic interactions with the surroundings, and outperforms baseline methods in both quantitative and qualitative aspects. 

\end{abstract}

\section{Introduction}
Recent advances in text-to-image (T2I) generative models  \cite{rombach2022high, ramesh2022hierarchical, saharia2022photorealistic} have enabled the rendition of highly realistic and creative images. These models are trained on large datasets of image-text pairs like LAION \cite{schuhmann2022laion}, allowing them to generate novel images from text prompts. In particular, there has been growing interest on T2I personalization \cite{gal2022image, ruiz2023dreambooth}. The T2I personalization is a task of generating variations of user-provided images in novel contexts by modifying aspects such as pose, action, color, and interactions between subjects. 

The studies on T2I personalization initially focused on the synthesis regarding a single subject given multiple images containing the subject \cite{gal2022image, ruiz2023dreambooth} by finetuning the text embedding representing the subject. Subsequent works \cite{wu2023singleinsert, wang2024stableidentity, hua2023dreamtuner}  explored personalization of a single subject based on a single image by using an additional image encoder. 
However, they had difficulties in varying the actions of personalized objects or required additional constraints like depth maps or skeletons. Recent works \cite{avrahami2023break,kim2024instantfamily, matsuda2024multi} proposed personalization of multiple subjects from a single image, enabling more convenient and diverse applications. However, they struggled to modulate diverse actions \cite{avrahami2023break} or needed explicit spatial guidance on subject regions when generating additional images \cite{kim2024instantfamily, matsuda2024multi}.

We aim to achieve multi-subject personalization from a single image while avoiding overfitting by introducing regularization through attention maps guided by the subject's concept-based prior. These priors, including class information, capture typical behaviors and enhance generalization. Unlike existing approaches that use MSE loss between segmentation masks and attention maps to isolate identities \cite{avrahami2023break, xiao2023fastcomposer, wang2024customvideo}, which ignore conceptual priors, our method employs loss functions on attention maps informed by subject concepts.

Furthermore, we implement concept-based prompt and image augmentation to encompass the subject's descriptive attributes and actions. Specifically, our Guided SDE Augmentation (GSA), inspired by SDEdit \cite{meng2021sdedit}, directs a T2I model with augmented prompts to generate images that balance identity preservation and action diversity. These augmentations enable diverse appearances, actions, and dynamic interactions among subjects. Experiments show that DynASyn produces images that are quantitatively and qualitatively superior to prior state-of-the-art methods.
Our main contributions are summarized as follows.
(1) We propose DynASyn for personalizing multiple subjects from a single image, which aligns the concept-based priors with the subject appearances and actions to mitigate overfitting. (2) We propose concept-based attention regularization and prompt-and-image augmentation for an effective alignment with concept priors. 
(3) Experiments on diverse datasets and prompts demonstrate that DynASyn achieves state-of-the-art personalization capabilities of synthesis for novel contexts and dynamic actions for the subjects. 


\begin{figure*}[t]
    \centering
    \includegraphics[width=\linewidth]{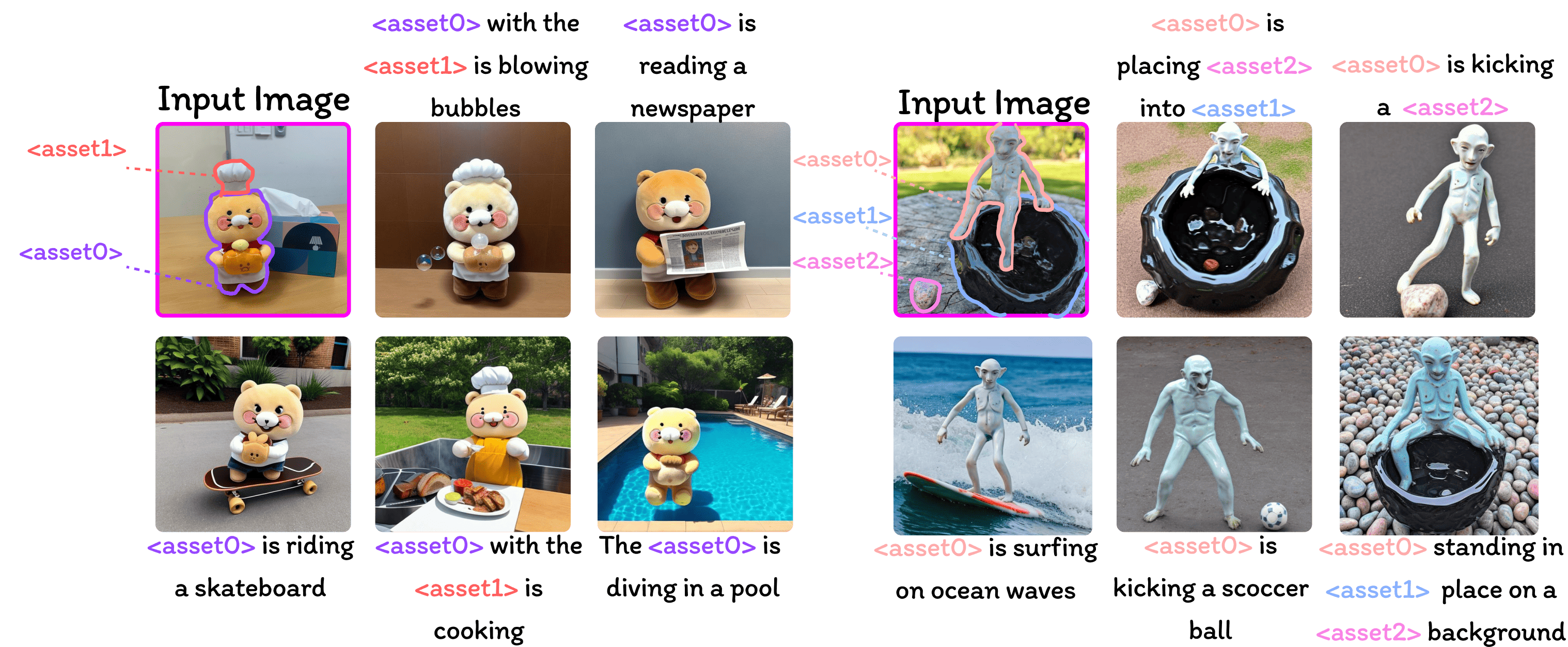}
    \caption{Personalization outputs from the proposed method, DynASyn. When provided with a single image containing multiple subjects, each subject can be trained into placeholders denoted as \texttt{<asset>}. DynASyn is capable of synthesizing diverse types of novel poses and dynamic actions of the subjects from the text prompts by avoiding overfitting to the reference image.}
    \label{fig:visualization}
\end{figure*}

\section{Related Work}
\textbf{Text-to-image Synthesis and Editing} The advent of large-scale image-text datasets \cite{schuhmann2022laion} and computing resources have spurred rapid innovations in text-to-image (T2I) synthesis models. Beginning with GAN based T2I model \cite{kang2023scaling, li2022stylet2i}, the field transitioned to transformer-based architectures \cite{ding2022cogview2, ramesh2021zero}. More recently, diffusion-based models \cite{rombach2022high, ramesh2022hierarchical, yu2022scaling, saharia2022photorealistic} have demonstrated remarkable progress. Diffusion models for T2I learn to progressively denoise images or latent representations from Gaussian noise, overcoming many limitations of existing  generative models based on GANs. When conditioned on text, they can produce strikingly realistic and diverse images. However, a major shortcoming is the lack of control over consistent subject generation or steering outputs towards desired target images.

Seeking more control beyond text-conditional image generation, recent work explores editing diffusion models by removing, inserting, or modifying parts of a given image. Some approaches edit the cross-attention map to alter sections of the image or replace subjects \cite{chefer2023attend, feng2022training}. Others incorporate additional signals like bounding boxes to control spatial regions \cite{chen2024training, li2023gligen}. Recently,  combining diffusion models with large language models (LLMs) in prompt-to-prompt frameworks \cite{hertz2022prompt} has enabled new pipeline designs for controllable editing.

\noindent\textbf{T2I Personalization} T2I personalization involves adapting models to generate images of a particular object given a few reference photos, conditioned on novel text prompts. Textual Inversion (TI) \cite{gal2022image} maps a set of 4-6 images of a subject to a placeholder token, training the text embedding to point to that specific subject. In TI, the small number of parameters updated during training can limit the identity preservation. DreamBooth (DB) \cite{ruiz2023dreambooth} follows a similar paradigm but also finetunes the parameters of the UNet to strengthen the identity preservation. However, DB can overfit to the reference images, which limits the generalization capabilities.
The research on simultaneous personalization of multiple subjects initially focused on human faces. Those works trained additional face encoders \cite{xiao2023fastcomposer, ruiz2023hyperdreambooth, wei2024mm} to disambiguate identities or used loss functions between segmentation masks and attention maps of different subjects \cite{xiao2023fastcomposer, wei2024mm}.  CustomDiffusion 
\cite{kumari2023multi} and Perfusion \cite{tewel2023key} enable multi-subject personalization from a few images by solely updating the key and value matrices in the cross-attention layer. Recent methods \cite{gal2023encoder, wei2023elite, jia2023taming, li2024blip} enable multi-subject personalization from a single image, but they require training additional encoders similar to the face personalization, which demands substantial computing resources and data. Break-a-Scene \cite{avrahami2023break} was able to personalize multiple subjects present in a single image. The method distinguished subjects using losses between segmentation masks and attention maps per subject, and trained place-holder tokens via an objective function combining TI and DB. However, the  attention loss between subject attention maps and masks in Break-a-Scene may cause  overfitting to fine-grained appearance details of the reference image. As a result, generations conditioned on new text tend to similar to the input image or to have difficulties in expressing novel poses or actions of the subject.


\section{Preliminaries}

\subsection{Text-to-Image Diffusion Models}

Text-to-image (T2I) models \cite{rombach2022high, ramesh2022hierarchical,saharia2022photorealistic} trained on large-scale image-text datasets synthesize high-fidelity images which accurately reflect the visual concepts expressed in textual descriptions. 
During training, the model learns to predict the text-conditional noise residual between the original and noisy images. For sampling, the model progressively denoises random images conditioned on the text embedding to generate the final image. Given a text encoder, $\gamma_\theta$, the training objective is given by
\begin{equation}\label{eqn: Diffusion Loss}
    \mathbb{E}_{z, c,\epsilon \sim \mathcal{N}(0,1),t} [||\epsilon - \epsilon_{\theta} (\gamma_\theta(c), z_t, t) ||^2_2]
\end{equation}
where $t$ is the timestep, $z_t$ is the  noisy latent, and $\epsilon_\theta$ is the denoising model. In T2I personalization, the model is trained on text prompts where a placeholder token denoted by \texttt{<asset>} represents the subject image. The model optimizes the embedding of placeholder tokens guided by the reconstruction loss. During training, text embeddings are incorporated as conditions via cross-attention with the spatial features of the UNet architecture. 
Specifically, the queries $Q$ come from an intermediate spatial latent of the UNet, and the keys $K$ and values $V$ are derived from the output of the text encoder network $\gamma_\theta(c)$. The resulting cross-attention map is \(
    A_t = \textrm{softmax}(QK^T/\sqrt{d})
\) where $d$ represents the dimension of the projected $K$ and $Q$. In $A_t \in \mathbb{R}^{m \times m \times N}$, $m$ is the spatial dimension of the attention map. $A_t$ plays a pivotal role in our method, and its use will be  explained in the sequel. We adopt Stable Diffusion v2.1 \cite{rombach2022high} as the main T2I model.

\subsection{Masked Diffusion}\label{sec:md}
Recent approaches to multi-subject personalization utilized the segmentation masks of subjects during training \cite{avrahami2023break, xiao2023fastcomposer, wei2024mm}. The  reconstruction loss is applied to the region activated by the segmentation mask of each subject. We also take this approach, where the subjects at each training step are randomly selected via \emph{union sampling} \cite{avrahami2023break} as follows. Suppose there are a total of $N$ subjects, and let $\mathcal I =\{1,\ldots,N\}$ denote the set of subject indices. A non-empty random subset $s\subseteq \mathcal I$ is sampled at each training step. Let $M_i$ denote the mask of subject $i$ and $M_s = \bigcup_{i\in s} M_{i}$ denote the union of the masks. The \emph{masked diffusion loss} $\mathcal{L}_{\text{MD}}$ is defined as
\begin{equation}
    \mathcal{L}_{\text{MD}} = \mathbb{E}_{z, s, \epsilon \sim \mathcal{N}(0,1), t} \left[ \| \epsilon \odot M_s - \epsilon_{\theta}(z_t, t, \gamma_\theta(c)) \odot M_s \|_2^2 \right] \tag{1}
\end{equation}

which is calculated by applying the subjects' masks to the noisy image. 
In addition, the masked loss can be applied to the cross-attention map between the placeholder token and images obtained through text-conditioning. For subject $i\in s$ and time $t$, let $A_{t,i}(\gamma,z)$ denote the cross-attention map between the $i$-th token embedding from $\gamma$ and the latent $z$ at $t$. The masked attention loss $\mathcal{L}_{\text{M2A}}$ is given by 
\begin{equation}\label{Eqn: M2A loss}
    \mathcal{L}_{\text{M2A}} = \mathbb{E}_{z,s,t} \left[\sum_{i \in s} \left\| A_{t,i} (\gamma_\theta(c), z_t) - M_i \right\|_2^2 \right].
\end{equation}
A benefit of masked diffusion with cross-attention is that, the identity of subject can be well-preserved by guiding the alignment between the subject placeholder and images through attention. However, at the same time, $\mathcal{L}_{\text{M2A}}$ encourages the model to attend uniformly to all areas within the mask. This can cause the model to overfit to mask outlines and geometries of subjects, which limits the generation of novel poses or actions for personalization. Thus, there exists a tension between preserving identity and overfitting to input seed images.

\begin{figure}[t!]
    \raggedright
    \includegraphics[width=0.46\textwidth]{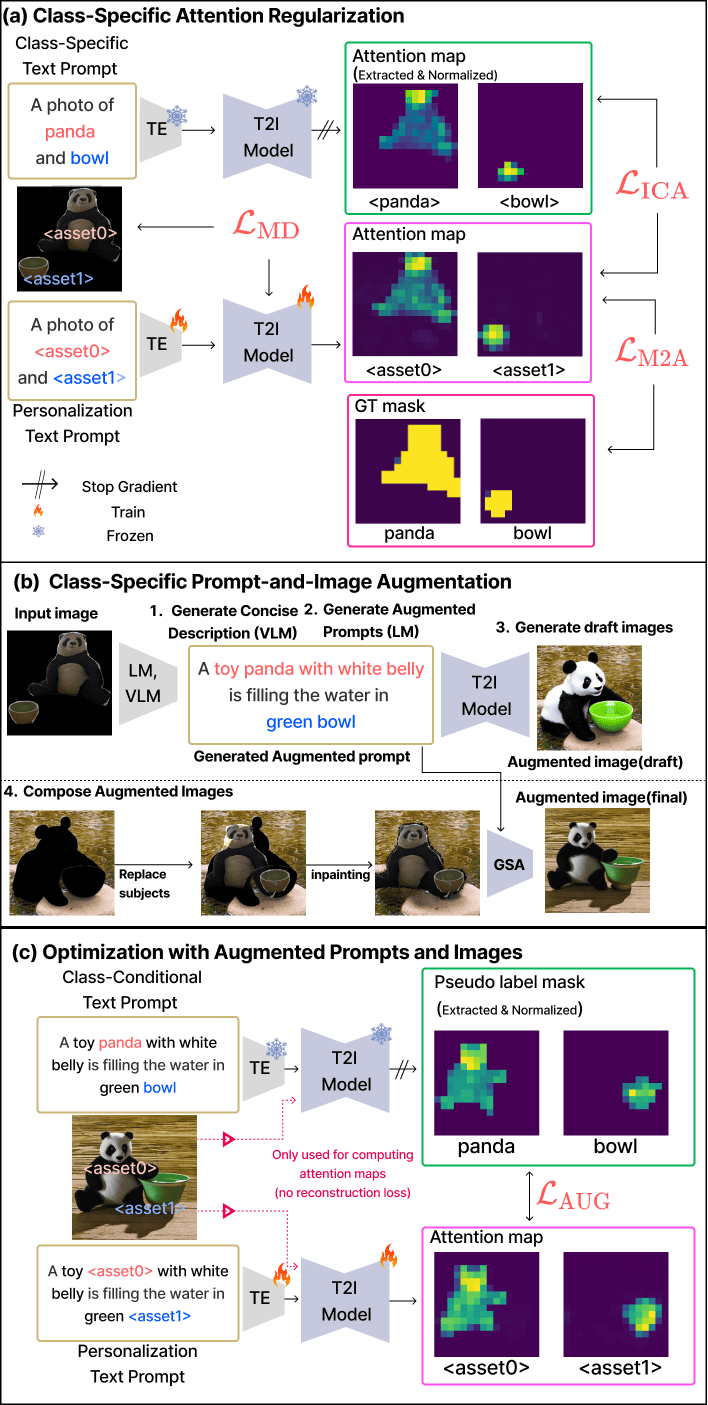}
  \caption{Overview of the DynASyn.
   (a) Concept-based Attention Regularization: the attention map derived from concept priors is used to regularize the attention map from token placeholder to prevent overfitting.
   (b) Concept-based Prompt-and-Image Augmentation: prompt-and-image pairs containing diverse action and poses of subjects are composed. 
   (c) Optimization with Augmented Prompts and Images: the augmented data from (b) is used for our model to learn to generate novel actions and poses of the subject.
   }
    \label{fig:overview}
\end{figure}

\begin{figure}[t!]
    \centering
    \includegraphics[width=0.45\textwidth]{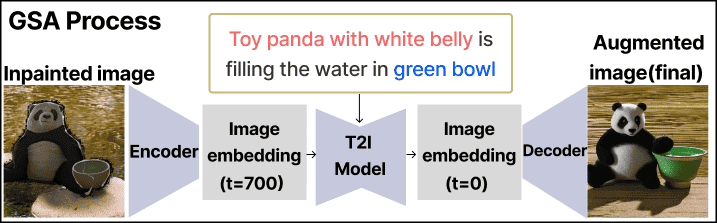}
    \caption{Overview of Guided SDE Augmentation (GSA).}
    \label{fig:sdedit}
\end{figure}

\section{Proposed Method} \label{method}
We aim to personalize \emph{multiple} subjects in a \emph{single} image by learning to disentagle the subjects and to generate novel concepts, actions  or interactions involving the subjects. A key to successful personalization is on how to properly integrate the subject identity and class-conditional priors of the subject. 
A vast knowledge of such priors can be found from large-scale T2I models, e.g., Stable Diffusion. The cross-attention map between the text describing the subject class and the associated images generated by T2I models contains rich information on the class-conditional image priors.
We propose to utilize those attention maps to infuse class-specific concepts into subject placeholders so as to facilitate generating novel appearances and actions consistent with the class priors.

\subsection{Class-Specific Attention Regularization} \label{sec:app}
We first focus on aligning subject \emph{appearances} with class-specific priors to mitigate overfitting issues. For optimizing the placeholder embeddings associated with the input images, a neutral prompt for the subject is used. For example, \url{a photo of <asset0> and <asset1>} is input to the text encoder, where \texttt{<asset0>} and \texttt{<asset1>} refer to the panda and bowl in the input image, e.g., see Fig.~\ref{fig:overview} (a). Next, we create another prompt with placeholders in personalization prompts replaced by class names of intended subjects. For example, in Fig.~\ref{fig:overview} (a), \url{a photo of <asset0> and <asset1>} becomes \url{a photo of panda and bowl}.

Thus, two prompts are used: one is the prompt with placeholders, and the other is the prompt with class tokens replacing placeholders. We then extract attention maps from the T2I model associated with two prompts. To regularize the attention between subject placeholders and input images, we introduce \emph{Inter-Cross-Attention (ICA)} loss. The ICA loss, denoted by $\mathcal{L}_{\text{ICA}}$, is defined as
{\small
\begin{equation}\label{Eqn: A2A loss}
    \mathbb{E}_{z,s,t} \left[ \sum_{i\in s} \left\| A_{t,i} (\gamma_\theta(c), z_t) - g\left( M_i \odot A_{t,i} (\gamma_\theta(\hat{c}), z_t) \right) \right\|_2^2 \right]
\end{equation}
}

where $c$ represents the sentence tokens with placeholders, $\hat{c}$ has placeholders replaced by class names, $\gamma_\theta$ is the text encoder, and $i$ indexes the placeholders. The function $g(x) = x/\max(x)$ normalizes the attention map. To suppress activations beyond subject boundaries in the attention map generated by the prompt with class tokens, we element-wise multiply the attention map by the masks and then normalize it. The attention maps of prompts with class tokens are obtained by frozen models with stop-gradient: see Fig. \ref{fig:overview}(a). Such attention maps regularize the cross-attention between subject placeholders and input images, enabling the model to capture both subject identity and class-specific priors.

\subsection{Class-Specific Prompt-and-Image Augmentation} \label{sec:act1}


Next, we focus on aligning the subject \emph{actions} with class-specific priors. While the ICA loss mitigates overfitting to subject appearances, it remains limited in freely generating subject actions such as novel poses or interactions among subjects. In order to generate flexible variations in actions, we propose concept-based prompt-and-image augmentation. The technique utilizes the vast prior knowledge about object concepts contained within large-scale models, such as large (vision) language models, T2I models, etc.

\textbf{Step 1: Generate Concise Description.} We query a pre-trained vision-language model (VLM) about the input image to personalize. The VLM is expected to provide a detailed description of the subjects in a single sentence. However, it is important to use a specific prompt that limits the number of words to encourage concise responses. We obtain a concise description of a subject in the form of a \emph{noun phrase}. Specifically, we input the subject image and the following prompt to the VLM:  ``\texttt{Tell me each subject in the photo less than 5 words using the noun phrase.}''. In the example of Fig. \ref{fig:overview}(b), the generated noun phrases associated with \texttt{<asset0>} (resp. \texttt{<asset1>}), denoted by \texttt{<P1>} (resp. \texttt{<P2>}), is \texttt{toy panda with white belly} (resp. \texttt{green bowl}). 



\textbf{Step 2: Generate Augmentation Prompts.} Based on concise descriptions of each subject from VLM, we generate multiple sentences on subject actions using a  language model (LM). The goal is to generate prompts describing inter-subject interactions or individual subjects performing diverse actions. The input to LM can be neutral sentences such as ``\texttt{Generate sentences of <P0> and <P1> interacting.}'' where \texttt{<P0>} and \texttt{<P1>} are replaced with the noun phrases generated in \textbf{Step 1}. The text box in Fig. \ref{fig:overview}(b) shows an example of generated prompt. This is the \emph{prompt augmentation}, and the augmented prompts will be used for training the generation model.



\textbf{Step 3: Generate Draft Images.} We input the augmented prompt to a T2I model and synthesize the associated images. In addition, pseudo-label masks for each subject are obtained by passing the rendered images through a generic, zero-shot segmentation model, e.g., Segment Anything Model (SAM) \cite{Kirillov_2023_ICCV}. 
The images generated in this step are based on general concepts (e.g., panda) and may not align well with the original input image for personalization. Thus, the generated image is used as a draft for composing augmented images in the next step.



\textbf{Step 4: Compose Augmented Images.} To better align draft images with the original subject identity, 
we perform a series of operations as follows:
see Fig. \ref{fig:overview}(b)-4. First, the subjects in draft images are replaced with the original version, ensuring spatial coherence. Then, any resulting gaps are filled using inpainting \cite{yu2023inpaint}.

Next, inspired by SDEdit \cite{meng2021sdedit}, we apply Guided SDE Augmentation (GSA), which is based on Stochastic Differential Equations (SDEs), to guide the T2I model with the augmented prompts. GSA starts with the inpainted image and performs 700 forward steps of diffusion. Then, it executes 700 backward steps using the T2I model conditioned on the augmented prompt from \textbf{Step 2} to obtain the final augmented image: see Fig. \ref{fig:sdedit}. GSA is specifically designed to align the augmented subject with the original through forward-backward SDE guided by augmented prompts (Fig. \ref{fig:sdedit}). This process ensures that the final augmented image is better aligned with the original subject compared to draft images, while still being capable of depicting various actions and poses through the guidance provided by the augmented prompts. The final augmented image is not a complete representation of personalization, but rather serves as an ``interpolation'' of diffusion sampling between the draft image and the inpainted image. For an example, compare the three images on the right of Fig. \ref{fig:overview}(b). In our implementation, we used GPT-4 \cite{achiam2023gpt} as VLM and LM, Stable Diffusion v2.1 \cite{rombach2022high} as the T2I model, and SAM \cite{Kirillov_2023_ICCV} as the segmentation model.

\subsection{Optimization with Augmented Prompts \& Images}\label{sec:act2}
Using the augmented prompts and images, we optimize the placeholder embeddings, text encoder and T2I model.  As previously, the cross-attention maps provided by the T2I model are utilized, where the attention maps between the augmented prompts and images are expected to capture various contextual cues and actions. We perform class-conditional attention regularization based on the augmented data as follows. Two prompts are used: one is the augmented prompt, and the other is the augmented prompt with the class token of the subject replaced by the placeholder token. Next, we extract attention maps from the T2I model using these sentences and augmented images, e.g., see Fig.\ \ref{fig:overview} (c). If  a reconstruction loss associated with input image is used, it can cause identity blending, because the model will be heavily influenced by individual pixel values from the augmented images. To maximize identity preservation while still allowing for action learning, we use only the inter-cross-attention loss without the reconstruction loss. The loss associated with the augmented data, denoted by $\mathcal {L}_{\text{AUG}}$, is defined as
\begin{equation}
    \mathbb{E}_{z,s,t} \left[ \sum_{i\in s} \left\| A_{t,i} (\gamma_\theta(c_a), z_t) - g(M_i \odot A_{t,i} (\gamma_\theta(\hat{c}_a), z_t)) \right\|_2^2 \right]
\end{equation}

where $c_a$ denote the text embeddings of the augmented prompt with placeholder,   $\hat{c}_a$ denote the embeddings of the same prompt with placeholders replaced by class tokens, and $M_i$ is the pseudo-label mask from the augmented image. 

The final training objective is given by
\begin{equation}\label{Eqn: total loss}
    \mathcal{L}  = \lambda_{\text{MD}}\mathcal{L}_{\text{MD}} + \lambda_{\text{M2A}}\mathcal{L}_{\text{M2A}} + \lambda_{\text{ICA}}\mathcal{L}_{\text{ICA}} + \lambda_{\text{AUG}}\mathcal{L}_{\text{AUG}}
\end{equation}
where  $\lambda_{\text{MD}}$, $\lambda_{\text{M2A}}$, $\lambda_{\text{ICA}}$, and $\lambda_{\text{AUG}}$ are hyperparameters for balancing the losses. By adjusting these hyperparameters, we strike a balance between identity regularization (e.g., $\mathcal{L}_{\text{MD}}$ and $\mathcal{L}_{\text{ICA}}$) and augmentation diversity (e.g., $\mathcal{L}_{\text{AUG}}$), ensuring that the model learns to preserve subject identity while generating diverse actions and interactions.

\begin{figure*}[!t]
    \centering
    \includegraphics[width=1.0\textwidth]{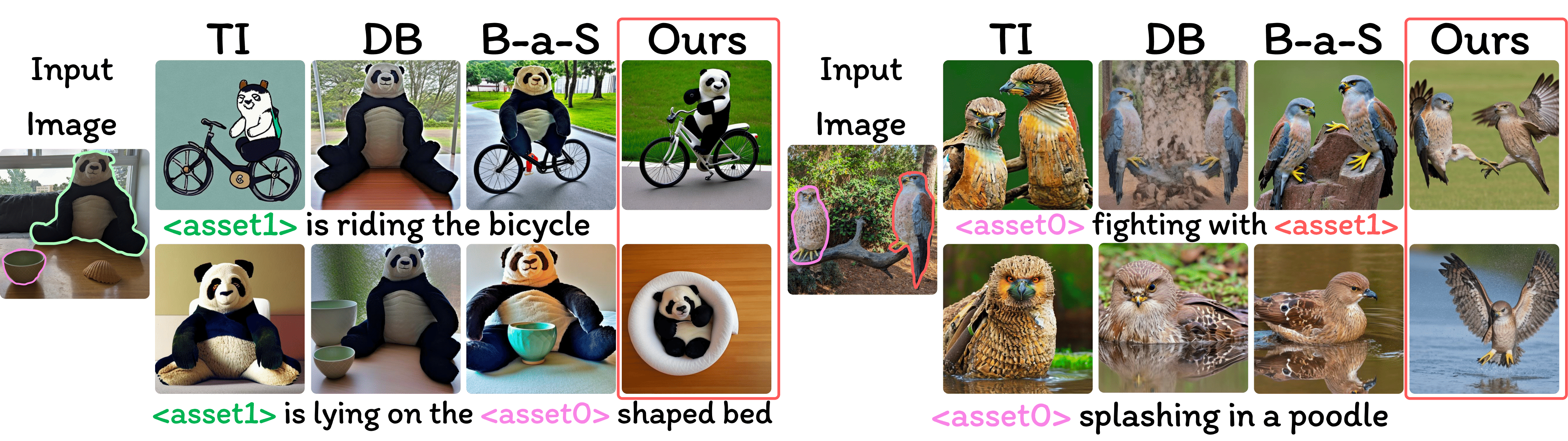}
    \caption{Qualitative comparisons with baseline methods. While baseline models often fail to align effectively with the provided text, DynASyn generates images that accurately reflect the textual input.}
    \label{fig:Comparison}
\end{figure*}

\begin{table*}[!t]
    \centering
    \begin{tabular}{c|ccc|ccc|ccc}
    \hline
                & \multicolumn{3}{c|}{Plain Sentence}                                                        & \multicolumn{3}{c|}{Action Sentence}                                           & \multicolumn{3}{c}{Interaction Sentence}                                       \\ \hline
                & \multicolumn{1}{c|}{$IR \uparrow$}         & \multicolumn{1}{c|}{$CLIP\text{-}T \uparrow$}         & $CLIP\text{-}I \uparrow$ & \multicolumn{1}{c|}{$IR \uparrow$}     & \multicolumn{1}{c|}{$CLIP\text{-}T \uparrow$}     & $CLIP\text{-}I \uparrow$ & \multicolumn{1}{c|}{$IR \uparrow$}     & \multicolumn{1}{c|}{$CLIP\text{-}T \uparrow$}     & $CLIP\text{-}I \uparrow$ \\ \hline
    TI (+mask)  & \multicolumn{1}{c|}{0.45}          & \multicolumn{1}{c|}{0.285}          & 0.71 & \multicolumn{1}{c|}{0.41}          & \multicolumn{1}{c|}{0.252}          & 0.688 & \multicolumn{1}{c|}{0.208}          & \multicolumn{1}{c|}{0.225}          & 0.659 \\ \hline
    DB (+mask) & \multicolumn{1}{c|}{0.713} & \multicolumn{1}{c|}{0.306}          & \textbf{0.84} & \multicolumn{1}{c|}{0.509} & \multicolumn{1}{c|}{0.274}          & \textbf{0.79} & \multicolumn{1}{c|}{0.459} & \multicolumn{1}{c|}{0.247}          & \textbf{0.823} \\ \hline
    B-a-S       & \multicolumn{1}{c|}{\underline{0.837}}           & \multicolumn{1}{c|}{\underline{0.351}}          & 0.774 & \multicolumn{1}{c|}{\underline{0.604}}          & \multicolumn{1}{c|}{\underline{0.31}}          & 0.732 & \multicolumn{1}{c|}{0.58}          & \multicolumn{1}{c|}{\underline{0.296}}          & 0.679 \\ \hline
    \textbf{DynASyn}     & \multicolumn{1}{c|}{\textbf{0.901}}          & \multicolumn{1}{c|}{\textbf{0.376}} & \underline{0.797} & \multicolumn{1}{c|}{\textbf{0.827}}          & \multicolumn{1}{c|}{\textbf{0.359}} & \underline{0.771} & \multicolumn{1}{c|}{\textbf{0.758}}          & \multicolumn{1}{c|}{\textbf{0.346}} & \underline{0.731} \\ \hline
    \end{tabular}
    \caption{Quantitative comparisons with baseline methods. \textbf{Plain sentences} are those with no significant change in action. \textbf{Action sentences} describe a change in behavior for a single subject. \textbf{Interaction sentences} involve multiple subjects interacting. Prompt fidelity (Image Reward Score (IR), CLIP-T), Subject fidelity (CLIP-I).}
    \label{Tab:main_result}
\end{table*}

\begin{table*}[!t]
\centering
\resizebox{\textwidth}{!}{
\begin{tabular}{c|ccc|ccc|ccc}
\hline
            & \multicolumn{3}{c|}{Plain Sentence} & \multicolumn{3}{c|}{Action Sentence} & \multicolumn{3}{c}{Interaction Sentence} \\ \hline
            & \multicolumn{1}{c|}{$Overall \uparrow$} & \multicolumn{1}{c|}{$Text \uparrow$} & $Identity \uparrow$ & \multicolumn{1}{c|}{$Overall \uparrow$} & \multicolumn{1}{c|}{$Text \uparrow$} & $Identity \uparrow$ & \multicolumn{1}{c|}{$Overall \uparrow$} & \multicolumn{1}{c|}{$Text \uparrow$} & $Identity \uparrow$ \\ \hline
TI (+mask)  & \multicolumn{1}{c|}{5.8} & \multicolumn{1}{c|}{6.6} & 4.7 & \multicolumn{1}{c|}{5.2} & \multicolumn{1}{c|}{4.7} & 6.3 & \multicolumn{1}{c|}{4.1} & \multicolumn{1}{c|}{2.3} & 4.3 \\ \hline
DB (+mask)  & \multicolumn{1}{c|}{3.7} & \multicolumn{1}{c|}{4.2} & \textbf{8.1} & \multicolumn{1}{c|}{5.5} & \multicolumn{1}{c|}{3.2} & \textbf{8.6} & \multicolumn{1}{c|}{5.6} & \multicolumn{1}{c|}{5.0} & \textbf{7.6} \\ \hline
B-a-S       & \multicolumn{1}{c|}{\underline{6.2}} & \multicolumn{1}{c|}{\underline{7.5}} & 6.6 & \multicolumn{1}{c|}{\underline{6.0}} & \multicolumn{1}{c|}{\underline{6.5}} & 6.3 & \multicolumn{1}{c|}{\underline{6.2}} & \multicolumn{1}{c|}{\underline{5.5}} & 6.1 \\ \hline
\textbf{DynASyn} & \multicolumn{1}{c|}{\textbf{8.2}} & \multicolumn{1}{c|}{\textbf{8.5}} & \underline{7.4} & \multicolumn{1}{c|}{\textbf{7.5}} & \multicolumn{1}{c|}{\textbf{7.4}} & \underline{7.5} & \multicolumn{1}{c|}{\textbf{7.7}} & \multicolumn{1}{c|}{\textbf{7.7}} & \underline{7.3} \\ \hline
\end{tabular}
}
\caption{Results of the user study. Identity measures how similar the subject in the generated image is to the original image, text evaluates how well the generated image reflects the textual input, and overall assesses the overall quality of the image.}
\label{Tab:user_study}
\end{table*}

\begin{figure*}[!t]
    \centering
    \includegraphics[width=1.0\textwidth]{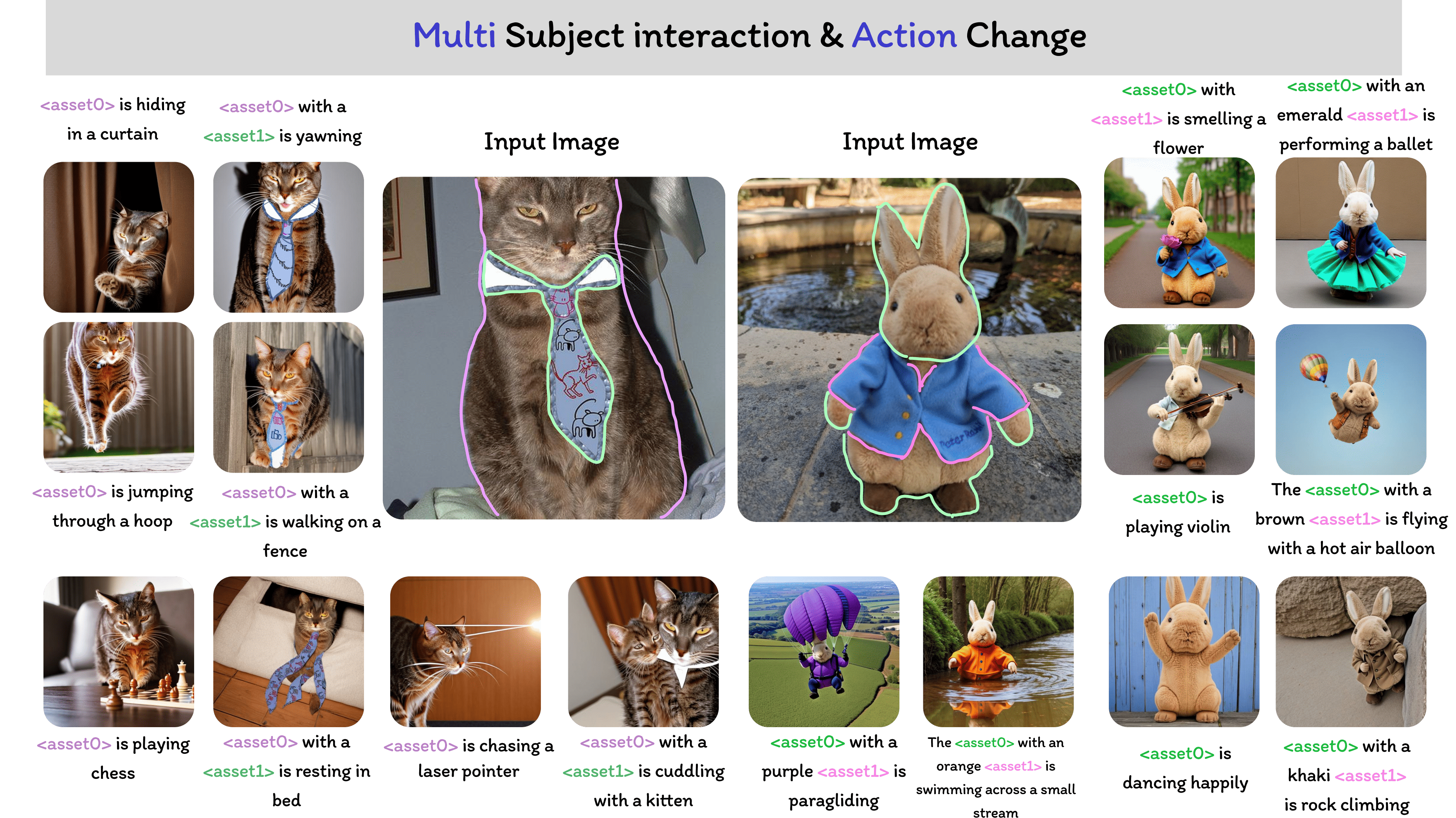}
    \caption{Visualization of personalized images. DynASyn generates a variety of images based on text when given a single input image. It can depict multiple subjects interacting dynamically or performing actions. Examples of additional generation tasks, such as re-contextualization or artistic stylization, can be found in Supplementary materials.}
    \label{fig:: visualization}
\end{figure*}

\section{Experiments}
\subsection{Experimental Setup} \label{sec. experimental_setup}
\sloppy
\textbf{Dataset.} We train and evaluate on a total of 15 image-text pairing datasets, including 7 individual datasets from Break-a-Scene, \\
2 datasets from the COCO benchmark \cite{lin2014microsoft}, \\
2 datasets from Zhang et al. \cite{zhang2024attention}, \\
and 4 additionally collected proprietary datasets. \\
All datasets contain at least two distinct subjects. \\
For the prompt-and-image augmentation, we generate 30 text prompts per image depicting new poses or inter-subject interactions using GPT-4 \cite{achiam2023gpt}.



\textbf{General Settings.} For all the methods, we utilized Stable Diffusion v2.1 \cite{rombach2022high}. 
We utilize Textual Inversion, Dreambooth, and Break-a-Scene as baselines for comparison. Since Textual Inversion and Dreambooth only allow single subject personalization, we use segmentation masks to restrict them to observing only the single subject per training step. Break-a-Scene enables multi-subject personalization, and thus we used it without modification. 
Detailed settings of hyperparmaeters are provided in Supplementary Materials.

\textbf{Evaluation Metrics.} 
We chose the metrics which gauge the similarity of generated images to the target subject and faithfulness to the conditioning text.
We assessed the subject similarity via CLIP-I scores and text alignment through both CLIP-T scores and Image Reward (IR) scores \cite{xu2024imagereward}. The IR score stems from a reward model trained on human judgments on text reflection and aesthetic quality for numerous text-image pairs.

\textbf{Results.} 
We conducted a comparison of the results for plain, action, and interaction by measuring text alignment and image similarity scores. Additionally, we performed a quantitative and qualitative comparison through user interviews obtained from 15 recruited volunteers.
As shown in Table 2, DynASyn achieves the highest text alignment, as measured by CLIP-T and image reward scores in all of the plain, action, and interaction sentences. 
As expected, TI (Textual Inversion) \cite{gal2022image} and DB (Dreambooth) \cite{ruiz2023dreambooth}, tend to be ineffective for multi-subject or single image personalization due to poor text reflection. Despite being able to personalize a single image, Break-a-Scene \cite{avrahami2023break} exhibits lower CLIP-T and image reward scores due to overfitting to appearances. In contrast, DynASyn resolves overfitting through prior-based personalization, yielding superior text alignment. DB attains the highest CLIP-I score since it is overfitted to the input, causing CLIP-I to favor output similarity, but at the cost of disregarded text conditions. Besides, DynASyn produces reasonable CLIP-I scores. 

This trend is also observed in the user interviews. DynASyn demonstrates superior performance, showing the highest overall and text alignment scores across all sentence types—Plain, Action, and Interaction—thereby validating its effectiveness. In contrast, TI and DB exhibit significantly lower overall and text alignment scores, with particularly poor performance on Action and Interaction sentences. While Dreambooth achieves the highest identity score, this is likely due to overfitting issues. Break-a-Scene shows lower text alignment and overall quality compared to DynASyn.
Qualitative results in Fig. \ref{fig:Comparison} show the comparison with the baseline methods\footnote{We 
attempted to make comparisons with recent works \cite{kumari2023multi} and \cite{zhang2024attention}. A crucial step in those works involves creating a regularization image set similar to the input image. The step relied on the LAION dataset \cite{schuhmann2022laion} which however was taken down since December 2023 and was unavailable at the time of this writing. Thus, we inspected the generated samples in \cite{kumari2023multi} and \cite{zhang2024attention} instead of a direct comparison. We find that, while both of these works personalize well, the personalization prompts primarily involved changing styles or scenes, or inserting new elements, but with few examples of altering poses or showing dynamic interactions with the surroundings. In contrast, our model is capable of generating a diverse range of actions by the subjects.}, which confirms the aforementioned trends. DynASyn personalizes subjects without overfitting and reflects text well. TI struggles with identity preservation. DB is overfitted to the input image, barely modifying outputs based on text. Break-a-Scene reflects text more poorly although less overfitting than DB. 

\begin{table}[]
\centering
\begin{tabular}{c|c|c|c|c|c}
\hline
MD & App & Act & $IR \uparrow$ & $CLIP\text{-}T \uparrow$ & $CLIP\text{-}I \uparrow$ \\ \hline
  $\,\checkmark$ &     &     & 0.341 & 0.292 & 0.767 \\ \hline
  $\,\checkmark$ & $\,\checkmark$ &     & 0.616 & 0.298 & \textbf{0.786} \\ \hline
  $\,\checkmark$ &     & $\,\checkmark$ & 0.732 & 0.306 & 0.76 \\ \hline
  $\,\checkmark$ & $\,\checkmark$ & $\,\checkmark$ & \textbf{0.829} & \textbf{0.360} & 0.767 \\ \hline
\end{tabular}
\captionof{table}{Ablation study (MD: $\mathcal{L}_{\text{MD}}, \mathcal{L}_{\text{M2A}}$; App: $\mathcal{L}_{\text{ICA}}$; Act: $\mathcal{L}_{\text{CSA}}$)}
\label{Tab:ablation}
\end{table}



\textbf{Ablation Study.} We conducted an ablation study to validate the efficacy of DynASyn components, comparing four model variants. The default model uses only masked diffusion (``MD'' in Table 3), while we toggle the alignment of subject appearances (Step (a) in Fig. \ref{fig:overview}, ``App'') and actions (Step (b), (c), ``Act''). As shown in Table 3, the CLIP-I scores show little overall difference across methods, which shows that overfitting was avoided while the identity is maintained. The CLIP-T scores, which measure the alignment of the text, also remain similar but improve slightly by adding the ``App'' component, which reduces the appearance overfitting. The inclusion of the ``Act'' component further enhances CLIP-T by learning various behaviors, with the full model achieving the highest CLIP-T and IR scores by balancing overfitting and diverse behavior learning.

\begin{figure}[!t]
  \centering
  \includegraphics[width=\linewidth]{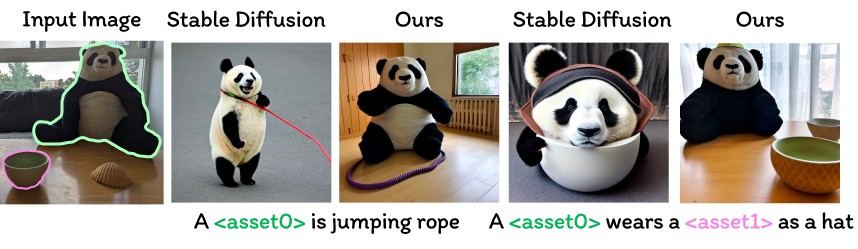} 
  \captionof{figure}{Limitation. In Stable Diffusion, We used class noun instead of placeholder \texttt{<asset>} for sampling.}
  \label{fig:limitation}
\end{figure}

\section{Conclusions and Limitations} \label{sec. limitation}
In this paper, we introduce DynASyn, a novel approach that leverages concept-specific regularization of cross-attention maps to effectively maintain identity representation while generating high-quality renditions across a wide array of prompt contexts.
To mitigate overfitting, we have developed a prompt-image augmentation technique that creates diverse prompt-image pairs encompassing a variety of actions and interactions, thereby enhancing the model's robustness and generalization capabilities.
Despite its promising performance, DynASyn presents certain limitations, as it is inherently dependent on prior knowledge and the performance of the underlying Text-to-Image (T2I) backbone.
As illustrated in Fig. \ref{fig:limitation}, models like Stable Diffusion struggle with challenging prompts, leading to a bottleneck where the accurate alignment of subject actions with concept priors is heavily reliant on the quality of generated images.
To address these issues, it is necessary to explore additional fine-tuning strategies that could further refine the model's adaptability.
Future work should focus on fine-tuning the T2I model using high-quality, expansive datasets, which will likely overcome the current limitations and facilitate more reliable and versatile image generation capabilities.
\clearpage
\section*{Acknowledgements}
This work was supported by the ICT Creative Consilience Program through the Institute of Information \& Communications Technology Planning \& Evaluation (IITP) grant funded by the Korea government (MSIT) (IITP-2024-2020-0-01819) and by the National Research Foundation of Korea (NRF) grant funded by the Korea government(MSIT)(RS-2022-NR070834).

\bibliography{aaai25}

\clearpage
\normalsize
\appendix
\section{Appendix}

\subsection{Training Details} \label{sup: experimental details}
This section covers the details for experiments, including hyperparameters. Training and sampling were performed using 1 A6000 GPU with 35GB GPU memory. The hyperparameters used in the experiments are listed in Table \ref{tab:hyper}.

\begin{table}[h!]
    \centering
    \begin{tabular}{|c|c|}
        \hline
        \textbf{Hyperparameter} & \textbf{Value}  \\
        \hline
        Training phrase 1 iteration & 700\\
        Training phrase 2 iteration & 700\\
        Phrase 1 Learning rate & $5\times 10^{-4}$\\
        Phrase 2 Learning rate & $2\times 10^{-6}$ \\
        $\lambda_{\text{MD}}$ & 1 \\
        $\lambda_{\text{M2A}}$ & $10^{-2}$ \\
        $\lambda_{\text{ICA}}$ & $5\times 10^{-3}$ \\
        $\lambda_{\text{AUG}}$& $1\times 10^{-4}$ \\
        Optimizer & Adam \\
       \hline
    \end{tabular}
    \caption{Detailed hyperparameters.} 
    \label{tab:hyper}
\end{table}
\begin{figure}[!h]
    \raggedright
    \includegraphics[width=0.5\textwidth]{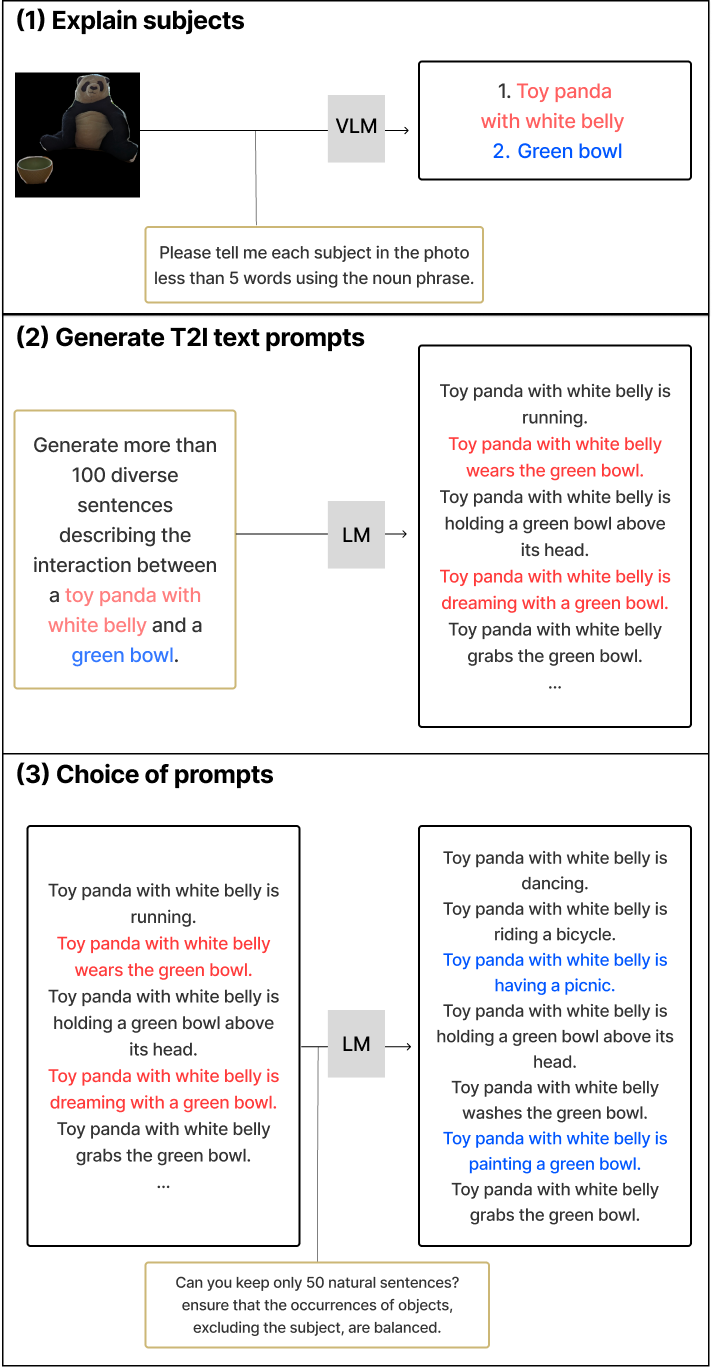}
    \caption{Process of prompts generation. (1) The image and query are provided to a VLM to extract the concepts of each subject. (2) Noun phrases are combined to generate diverse sentences using a LM. Red highlights indicate unnatural sentences. (3) Unnatural sentences are filtered out by the LM, and the remaining sentences are reviewed, with 30 to 50 of the most natural ones selected for model training. Red highlights show removed sentences; blue and black show selected sentences.}
    \label{fig:user_interview}
\end{figure}








\subsection{Process of prompts generation} \label{sup: prompt_generation_1}
This section details the specific process and prompts used in train and inference process. First, the image is provided as context to a Vision-Language Model (VLM) along with the query: ``Please describe each subject in the photo in less than 5 words using a noun phrase.'' The VLM then is expected to provide a concise noun phrase for each subject.

Next, these noun phrases are combined to generate diverse sentences using a Language Model (LM). In this process, the following text query is used: 
``Generate more than 100 diverse sentences describing the interaction between a toy panda with a white belly and a green bowl.'' At this step, somewhat awkward or unnatural sentences might be included. Therefore, we additionally use the following prompt to eliminate any awkward sentences: ``Can you keep only 50 natural sentences and ensure that the occurrences of objects, excluding the subject, are balanced?'' From the remaining sentences, between 30 to 50 of the most natural sentences are selected for training.

\subsection{Categorized Prompts for Evaluation}

A similar process for training was utilized for generating prompts for inference. In our evaluation, we categorized the personalization prompts into three categories:
\begin{itemize}
    \item Plain: relatively static changes in terms of subjects' motions, e.g., changes in poses, appearance, style, etc.
    \item Action: a single subject performing dynamic actions.
    \item Interaction: dynamic interaction between multiple subjects.
\end{itemize}
Below are examples of 10 sentences used in each category.  Here, \texttt{<asset0>} refers to \textit{a toy panda with a white belly}, and \texttt{<asset1>} refers to \textit{a green bowl}.

\subsubsection*{Plain}
\begin{itemize}
    \item \texttt{<asset0>} in a nurse suit.
    \item \texttt{<asset0>} in an Iron Man suit.
    \item \texttt{<asset0>} in a firefighter uniform.
    \item \texttt{<asset0>} in a pencil sketch.
    \item \texttt{<asset0>} in an oil painting.
    \item \texttt{<asset0>} in a comic.
    \item \texttt{<asset0>} in watercolor.
    \item \texttt{<asset0>} at the Acropolis.
    \item \texttt{<asset0>} at the Eiffel Tower.
    \item \texttt{<asset0>} in a jungle.
\end{itemize}
\begin{figure}[!h]
    \raggedright
    \includegraphics[width=0.5\textwidth]{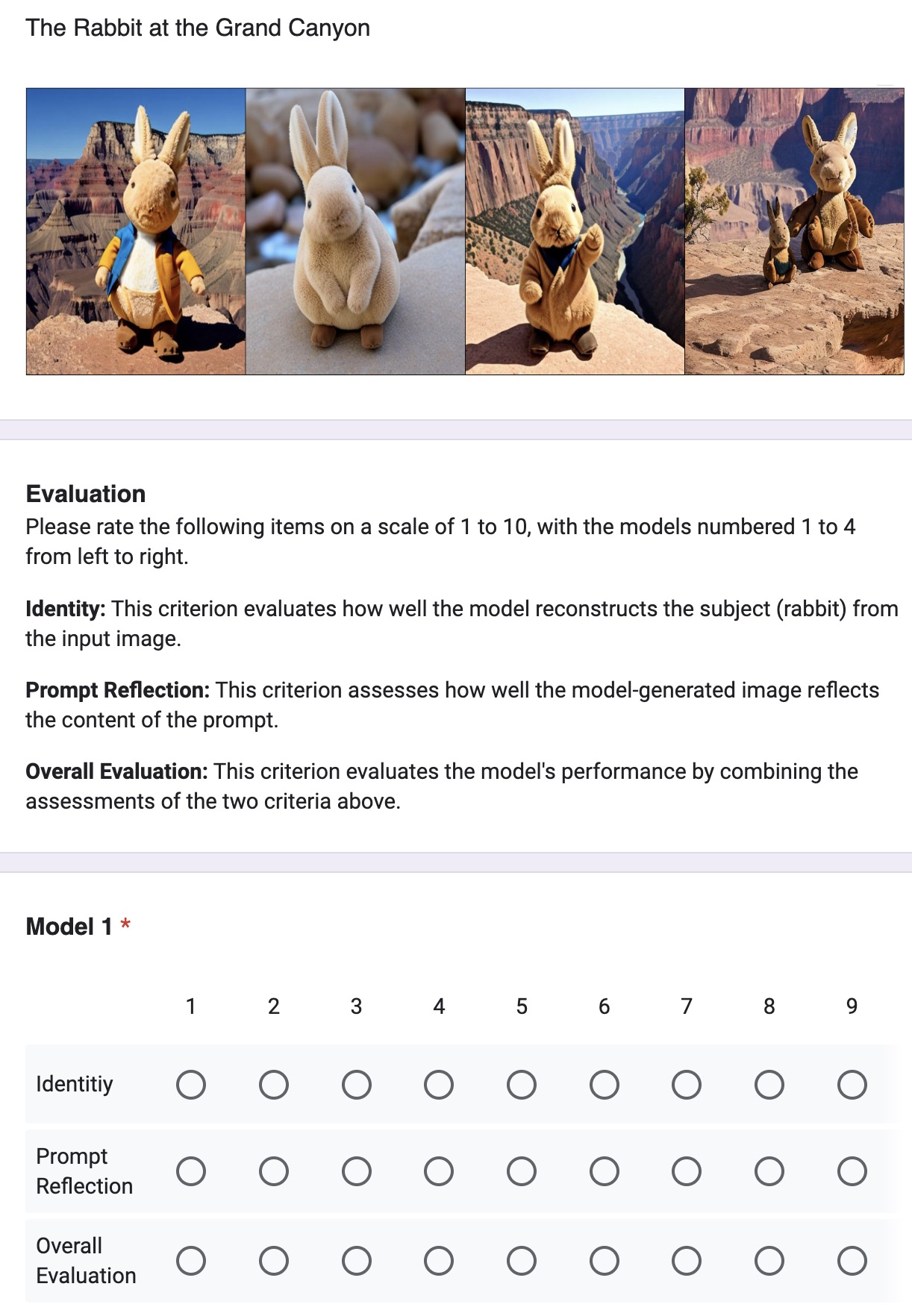}
    \caption{Screenshot of some user interviews. From left to right, the images were generated by Break-a-Scene, Dreambooth, DynaASyn, and Text Inversion.}
    \label{fig:user_interview}
\end{figure}
\subsubsection*{Action}
\begin{itemize}
    \item The \texttt{<asset0>} is running.
    \item The \texttt{<asset0>} is jumping.
    \item The \texttt{<asset0>} is sitting.
    \item The \texttt{<asset0>} is dancing.
    \item The \texttt{<asset0>} is sleeping.
    \item The \texttt{<asset0>} is reading a book.
    \item The \texttt{<asset0>} is playing a guitar.
    \item The \texttt{<asset0>} is eating a carrot.
    \item The \texttt{<asset0>} is riding a bicycle.
    \item The \texttt{<asset0>} is flying a kite.
\end{itemize}

\subsubsection*{Interaction}
\begin{itemize}
    \item \texttt{<asset0>} is throwing the \texttt{<asset1>}.
    \item \texttt{<asset0>} is washing the \texttt{<asset1>}.
    \item \texttt{<asset0>} is filling the \texttt{<asset1>} with bamboo leaves.
    \item \texttt{<asset0>} is drinking water from the \texttt{<asset1>}.
    \item \texttt{<asset0>} is placing the \texttt{<asset1>} on a table.
    \item \texttt{<asset0>} is looking into the \texttt{<asset1>}.
    \item \texttt{<asset0>} is carrying the \texttt{<asset1>}.
    \item \texttt{<asset0>} is stacking the \texttt{<asset1>}s.
    \item \texttt{<asset0>} is painting the \texttt{<asset1>}.
    \item \texttt{<asset0>} is balancing the \texttt{<asset1>} on its head.
\end{itemize}
In our implementation, we used GPT-4 \cite{achiam2023gpt} as both the VLM and LM.

\subsection{Details of user interviews} \label{sup: data_distribution}
The user interviews presented in Table 2 were conducted for four models: DynASyn, Break-a-Scene, Dreambooth, and Text Inversion. We categorized the sentences into three major categories: Plain, Action, and Interaction. For each category, the evaluation was conducted using five samples. After training the models with the same samples, images were generated using identical sentences. The users were asked to rate them on a scale of 1 to 10. The evaluation criteria were the similarity of the generated images to the target subject, faithfulness to the conditioning text, and an overall score that combined these factors. An example of the user interview results is shown  in Fig.\ref{fig:user_interview}.


\subsection{Additional results of proposed method}\label{sup: additional_result}
\begin{figure*}[!h]
    \centering
    \includegraphics[width=0.8\textwidth]{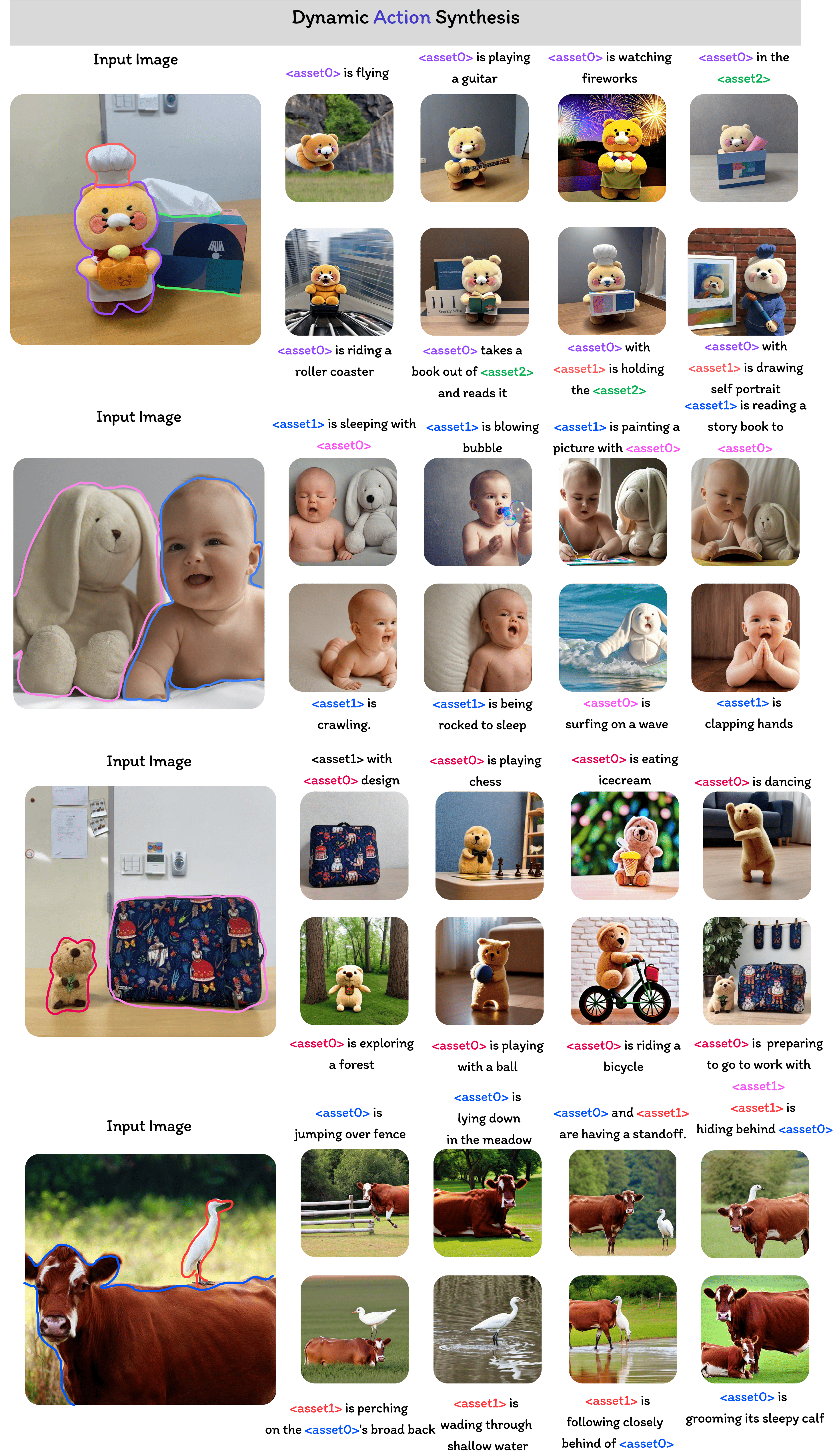}
    \caption{Additional Qualitative results. Dynamic Action Synthesis.}
    \label{fig:sup_vis1}
\end{figure*}

\begin{figure*}
    \centering
    \includegraphics[width=0.7\textwidth]{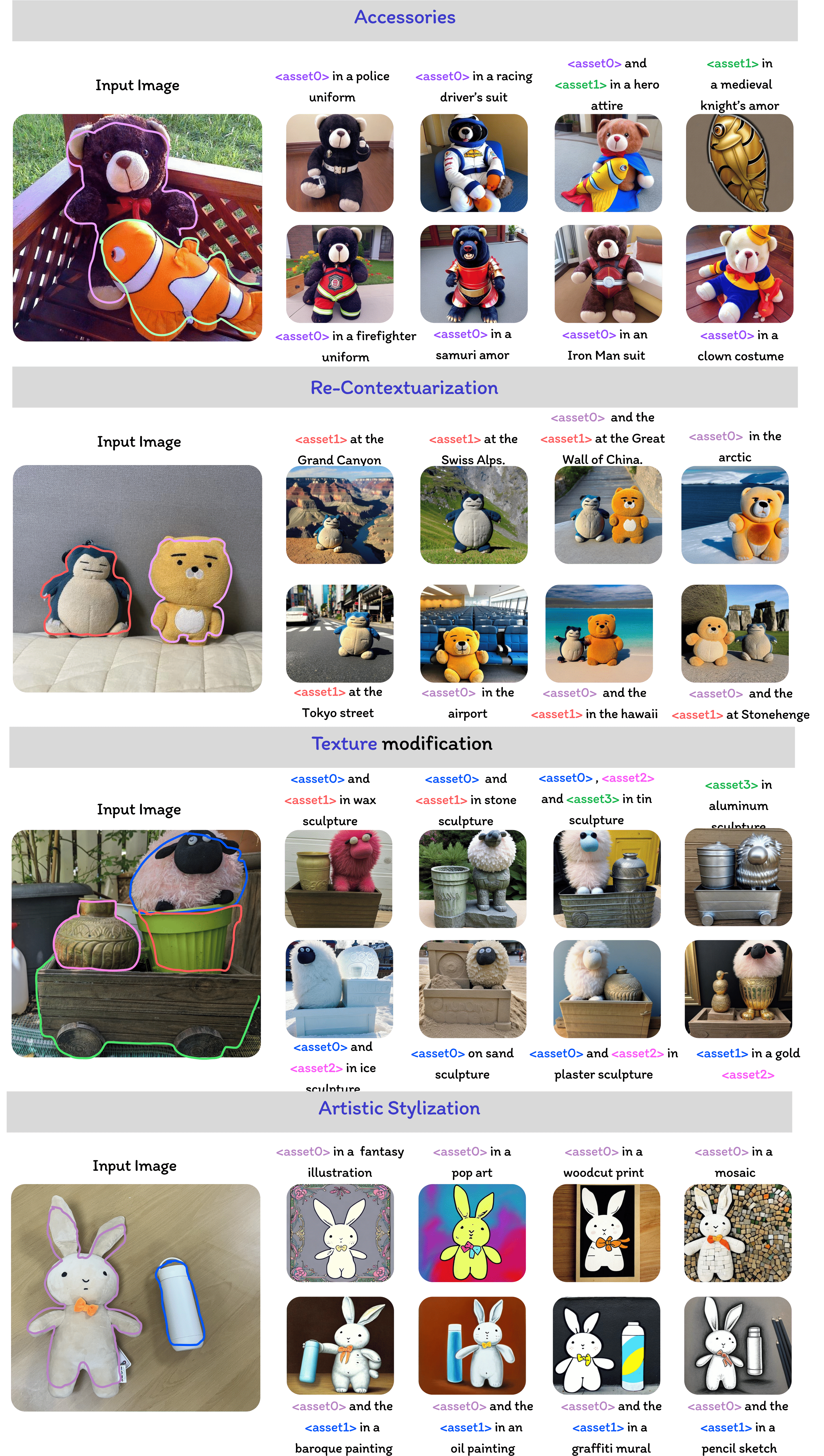}
    \caption{Additional Qualitative results. Accessories, Re-Contextuarization, Textual Modification, and Artistic Stylization.}
    \label{fig:sup_vis2}
\end{figure*}

In this section, we provide additional personalization results by DynASyn.
\begin{itemize}
\item Fig. \ref{fig:sup_vis1} presents  results on dynamic action synthesis.
\item Fig. \ref{fig:sup_vis2} presents results on changing accessories, re-contextuarization, textual modification and artistic stylization.

\end{itemize}

\subsection{Distributing the dataset} \label{sup: data_distribution}
\begin{figure*}
    \centering
    \includegraphics[width=0.9\textwidth]{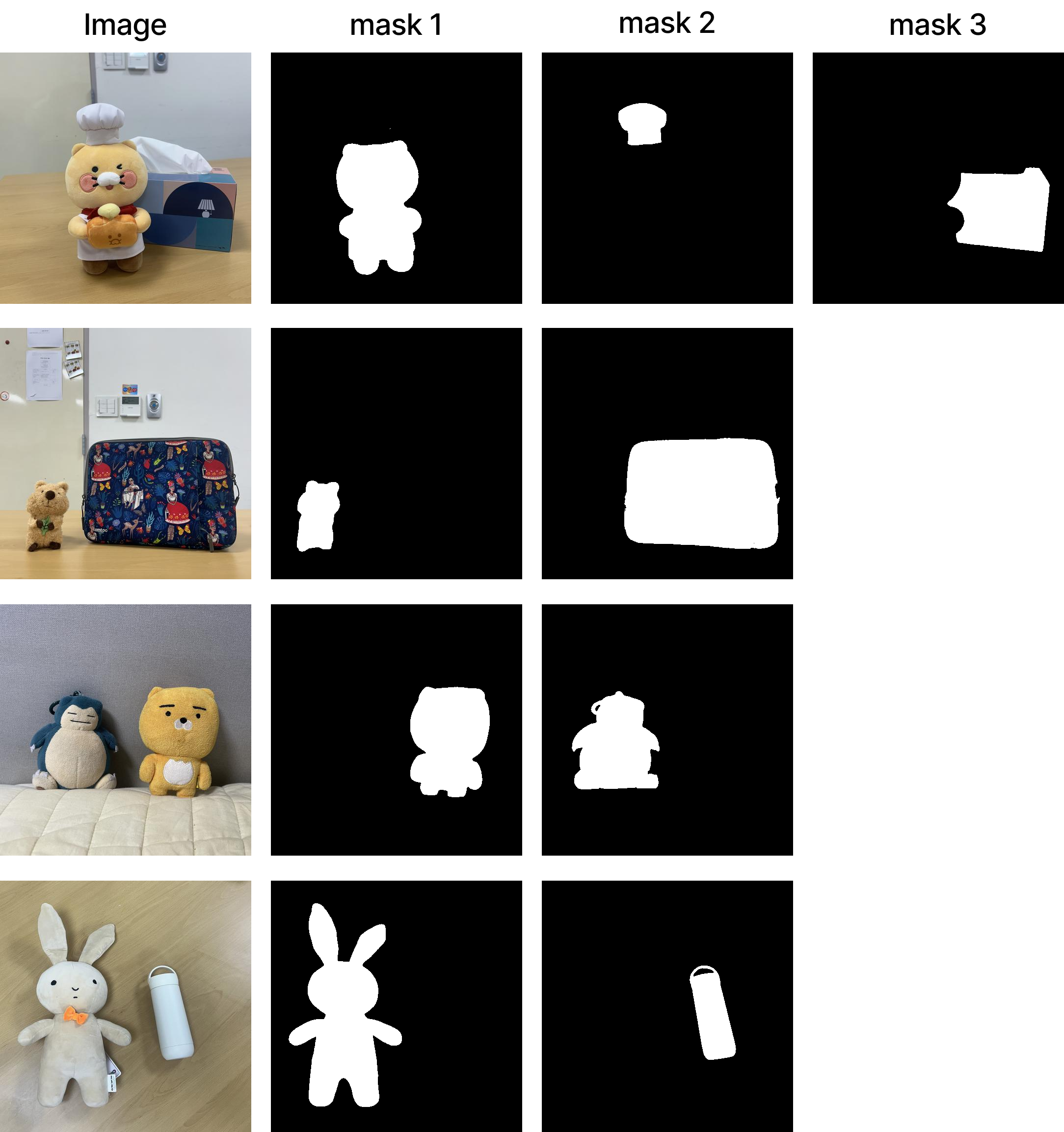}
    \caption{Custom data we collected for experiments.}
    \label{fig:dataset}
\end{figure*}

We release some samples of image-mask pairs for multi-subject personalization. The dataset includes various dolls and objects with each data sample containing 2 or more subjects. The images and subject masks in the dataset are shown  in Fig. \ref{fig:dataset}.

\end{document}